\documentclass[sigconf]{acmart}

\AtBeginDocument{%
  }

\setcopyright{acmlicensed}
\copyrightyear{2025}
\acmYear{2025}
\acmDOI{XXXXXXX.XXXXXXX}

\acmConference[Conference acronym 'XX]{Make sure to enter the correct conference title from your rights confirmation email}{June 03--05,
  2018}{Woodstock, NY}

\acmISBN{978-1-4503-XXXX-X/18/06}

\usepackage{multirow}
\usepackage{makecell}
\usepackage{pifont}
\usepackage[graphicx]{realboxes}
\usepackage[normalem]{ulem}
\useunder{\uline}{\ul}{}

\copyrightyear{2025}
\acmYear{2025}
\setcopyright{cc}
\setcctype{by-sa}
\acmConference[KDD '25]{Proceedings of the 31st ACM SIGKDD Conference on Knowledge Discovery and Data Mining V.2}{August 3--7, 2025}{Toronto, ON, Canada}
\acmBooktitle{Proceedings of the 31st ACM SIGKDD Conference on Knowledge Discovery and Data Mining V.2 (KDD '25), August 3--7, 2025, Toronto, ON, Canada}
\acmDOI{10.1145/3711896.3737440}
\acmISBN{979-8-4007-1454-2/2025/08}

\begin{document}

\title{SatHealth: A Multimodal Public Health Dataset with Satellite-based Environmental Factors}

\author{Yuanlong Wang}
\orcid{}
\email{wang.16050@osu.edu}
\affiliation{%
  \institution{The Ohio State University}
  \city{Columbus}
  \state{Ohio}
  \country{USA}
}

\author{Pengqi Wang}
\orcid{}
\email{wang.19883@osu.edu}
\affiliation{%
  \institution{The Ohio State University}
  \city{Columbus}
  \state{Ohio}
  \country{USA}
}

\author{Changchang Yin}
\orcid{}
\email{yin.731@osu.edu}
\affiliation{%
  \institution{The Ohio State University}
  \city{Columbus}
  \state{Ohio}
  \country{USA}
}

\author{Ping Zhang}
\orcid{https://orcid.org/0000-0002-4601-0779}
\email{zhang.10631@osu.edu}
\authornote{Corresponding Author}
\affiliation{%
  \institution{The Ohio State University}
  \city{Columbus}
  \state{Ohio}
  \country{USA}
}

\renewcommand{\shortauthors}{Wang et al.}
\newcommand{\yin}[1]{\textcolor{blue}{#1}}
\newcommand{\zhang}[1]{\textcolor{green}{#1}}
\newcommand{\wang}[1]{\textcolor{red}{#1}}
\newcommand{\cmark}{\ding{51}}%
\newcommand{\xmark}{\ding{55}}%
\newcommand{\eg}{\textit{e.g.}, }
\newcommand{\ie}{\textit{i.e.}, }


\begin{abstract}
Living environments play a vital role in the prevalence and progression of diseases, and understanding their impact on patient's health status becomes increasingly crucial for developing AI models.
However, due to the lack of long-term and fine-grained spatial and temporal data in public and population health studies, most existing studies fail to incorporate environmental data, limiting the models' performance and real-world application.
To address this shortage, we developed SatHealth, a novel dataset combining multimodal spatiotemporal data, including environmental data, satellite images, all-disease prevalences estimated from medical claims, and social determinants of health (SDoH) indicators.
We conducted experiments under two use cases with SatHealth: regional public health modeling and personal disease risk prediction. Experimental results show that living environmental information can significantly improve AI models' performance and temporal-spatial generalizability on various tasks. 
Finally, we deploy a web-based application\footnote{\label{web} \url{https://aimed-sathealth.net}} to provide an exploration tool for SatHealth and one-click access to both our data and regional environmental embedding to facilitate plug-and-play utilization. SatHealth is now published with data in Ohio, and we will keep updating SatHealth to cover the other parts of the US. 
With the web application and published code pipeline\footnote{\label{github} \url{https://github.com/Wang-Yuanlong/SatHealth}}, our work provides valuable angles and resources to include environmental data in healthcare research and establishes a foundational framework for future research in environmental health informatics. 
\end{abstract}




\maketitle
\section{Introduction}

\begin{figure*}[ht]
    \centering
    \includegraphics[width=0.8\linewidth]{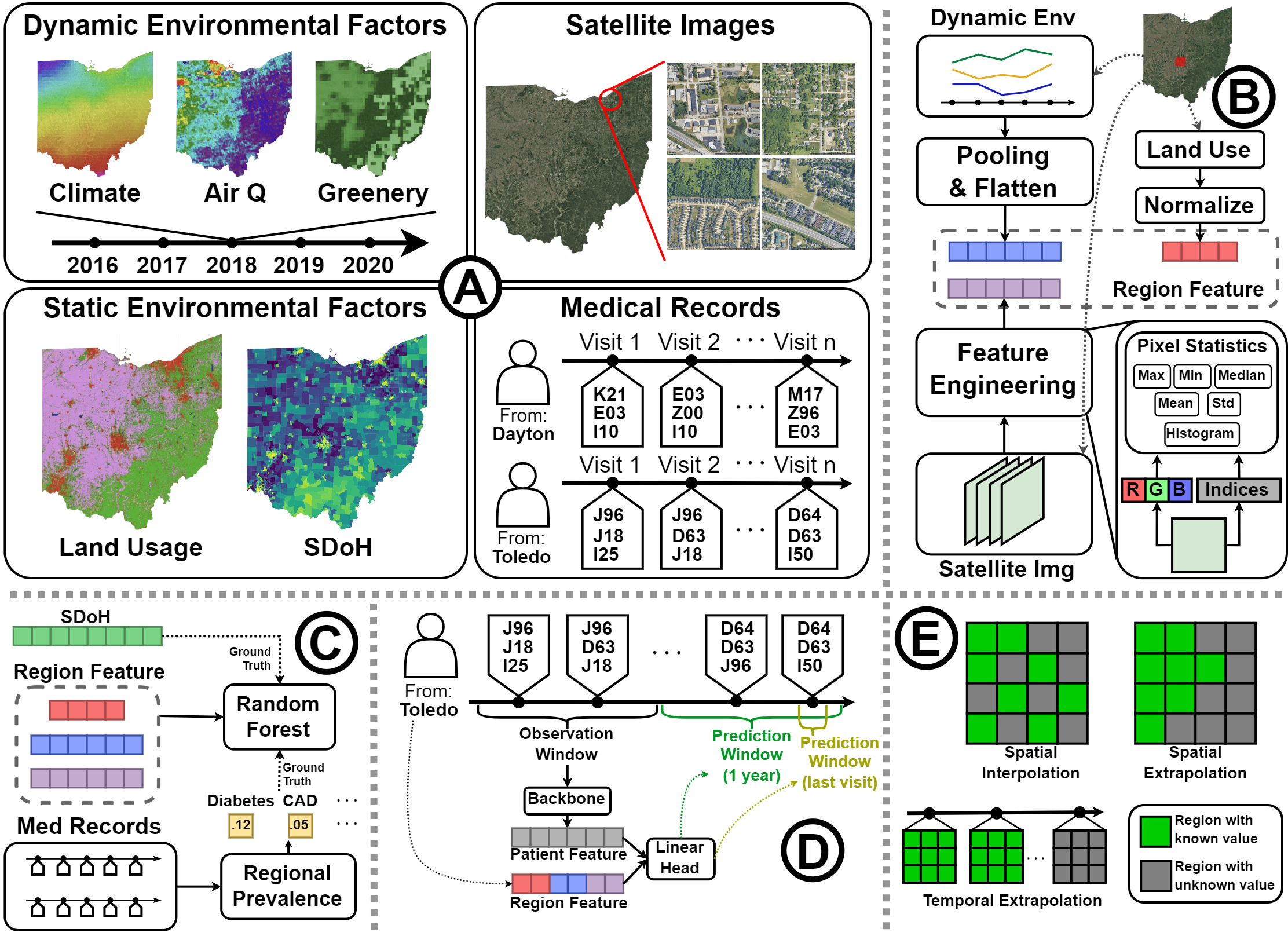}
    \vspace{-4mm}
    \caption{Overview of SatHealth and Experimental framework. (A) The components of SatHealth. (B) The regional embedding pipeline. We use all dynamic factors, satellite images, and land usage as the multimodal environmental features of a region. (C) We use the regional embedding to predict the prevalence of all kinds of diseases and SDoH within that region. (D) We combine patient representation with the regional embedding of patient residence to predict patient disease risk. (E) We test model robustness on three generalization scenarios.}
    \vspace{-3mm}
    \label{fig:data_overview}
\end{figure*}
 
The living environment (\eg climate, green spaces, air quality, and socioeconomic factors) considerably impacts people's physical~\cite{dadvand_inequality_2014, gascon_residential_2016} and mental~\cite{white_would_2013} health, and understanding such impact is an emerging topic~\cite{cleary_changes_2019, iamtrakul_association_2021, li_urban_2023}.
Considerable efforts have been made to explore how these factors are associated with human health~\cite{keenan_intersecting_2024, wafula_socioeconomic_2024, spoerri_swiss_2010}.
However, due to the complexity of aligning multi-source geospatial data with medical data, there are still limited handy medical datasets equipped with high spatiotemporal resolution, long-term coverage, and comprehensive environmental data. Therefore, existing studies either focus on specific diseases and environmental factors~\cite{luo_differing_2024, chen_artificial_2024} or take solely patients' medical record histories without consideration of environmental factors~\cite{yu_predict_2024, zang_identification_2024, chen_predictive_2024, xu_flexcare_2024}. As a result, these models might be limited in their accuracy, comprehensiveness, and spatiotemporal robustness. Moreover, the benefit of information from the living environment in healthcare AI remains unexplored.


To fill the shortage, we propose a pipeline and develop SatHealth, a compound dataset featuring satellite-based environmental data, satellite imagery, prevalence of all diseases estimated from medical claims, and Social Determinants of Health (SDoH) indices. To the best of our knowledge, SatHealth is the first dataset in the US that combines regional environmental characteristics with a healthcare database. 
We requested over 400k aerial view satellite images from Google Maps~\cite{google_static_map}, each image covers about a 500m wide square area. Furthermore, we use medical claims from the MarketScan~\cite{merative_real_2024} database to estimate the regional prevalence for all diseases. As for SDoH, we use the Social Deprivation Index (SDI)~\cite{sdi_def}, which is a comprehensive score calculated from US census data from the American Community Survey (ACS)~\cite{acs_def}. 
Moreover, we designed a multimodal fusion framework to seamlessly integrate heterogeneous multimodal environmental data sources and provide user-friendly regional environment embeddings, facilitating downstream analyses and follow-up studies.


We first validated and quantified the environmental-disease relationship by statistical testing, reflecting the disparity in urban-rural health status. 
After that, we use the dataset on two clinical tasks: regional public health modeling (\eg to predict regional SDI scores and disease prevalence based on environmental data) and personalized disease risk prediction (\eg to enhance personal disease risk prediction with environmental data). The experimental results show that living environmental information can significantly improve AI models’ performance and spatiotemporal generalizability.
Finally, we deployed a web-based application\textsuperscript{\ref{web}} where users could explore and access SatHealth data with regional embedding vectors. Our regional embeddings can be plugged into any clinical AI with geospatial information, which paves the way for integrating environmental factors into clinical AI development.

We started SatHealth development from Ohio as a concomitant of the Ohio O-SUDDEn program~\cite{osudden_proj}. However, all satellite data we use have global coverage, and MarketScan patient-level medical claims have US coverage. Therefore, our framework's environmental factor processing pipeline can be easily adapted to other areas. We also provide our code on GitHub\textsuperscript{\ref{github}} so that users can create data and embeddings for different areas of interest. We will also gradually update SatHealth to cover the US in the future.

\begin{table*}
\centering
\caption{A comparison of SatHealth to related datasets combining environmental and health data}
\vspace{-4mm}
\label{tab:related_works}
    \resizebox{0.98\linewidth}{!}{
    \begin{tabular}{lllllll}
    \toprule
Dataset        & Location                      & Target                                        & Target data source               & Imagery              & Year span & Public \\\midrule
Barboza et al. & 31 European countries         & Mortality                                     & Eurostat                         &    \xmark           & 2015      &  \xmark \\
Temenos et al. & 8 European cities             & COVID-19                                      & OWD &   \xmark             & 2020-2021 & \xmark \\
SatelliteBench & 81 municipalities in Colombia & Dengue outbreak, Poverty, Access to school    & SIVIGILA, Census     & Sentinel-2           & 2016-2018 & \cmark \\
SustainBench   & Global                        & BMI, child mortality,water quality,sanitation & Surveys                          & LandSat, Street view & 1996-2019 & \cmark  \\
MedSat         & England                       & Medical prescription of 7 conditions          & NHS   & Sentinel-2           & 2019-2020 & \cmark  \\
\textbf{SatHealth}      & Ohio, US                      & SDoH, All ICD code prevalence                 & SDI, MarketScan                  & Google Maps           & 2016-2022 & \cmark \\\bottomrule
\multicolumn{7}{l}{OWD: Our World in Data platform; NHS: National Health Services; SIVIGILA: Colombian Public Health Surveillance System}\\  
\end{tabular}
}
\end{table*}

In summary, we summarize our contributions as follows:
\begin{itemize}
    \item We construct SatHealth, the first publicly available dataset in the US consisting of environmental data, satellite imagery, regional SDoH, and all-disease prevalence for comprehensive environment-health analysis.
    \item We design an embedding pipeline to construct regional environment embeddings by fusing multimodal data from SatHealth. 
    \item We show two use cases of SatHealth. The experimental results exhibit the benefit of environmental information in model accuracy and temporal-spatial generalizability. 
    \item We deployed a web-based application to showcase and provide access to SatHealth and our embeddings. We also publish our code for data collection in other areas of interest.
\end{itemize}

\textbf{Code and Docs}:  \url{https://github.com/Wang-Yuanlong/SatHealth}

\textbf{Web app and Dataset download}: \url{https://aimed-sathealth.net}

\textbf{License}: The dataset is released under the CC BY-SA 4.0 license.

\section{Related Works}

\subsection{Environment-health Datasets}

Environmental factors, such as built environment, air quality, and green space, are reported to be correlated with many diseases, such as heart diseases~\cite{chen_deep_2024, chen_artificial_2024}, metabolic syndrome~\cite{lam_built_2023}, and stroke~\cite{asri_global_2020}. Researchers construct datasets with earth observation (EO) and remote sensing data with various health outcomes to learn their health impacts. For example, Barboza et al.~\cite{barboza_green_2021} and Temenos et al.~\cite{temenos_novel_2022} collected regional greenery indices like normalized difference vegetation index (NDVI) and percentage of green area (\%GA), and studied their relationship with natural-cause mortality and COVID-19 cases. However, the greenery indices provide a limited expression of the living environment. Therefore, some datasets include satellite images to capture a more comprehensive neighborhood view.  SatelliteBench~\cite{moukheiber_multimodal_2024} examines the dengue outbreak in 81 Colombian municipalities, collecting Sentinel-2 satellite images, climate, socioeconomic factors, and dengue cases to predict poverty, school access, and dengue outbreaks.
Moreover, SustainBench~\cite{yeh_sustainbench_2021} benchmarks prediction tasks of 4 public health indicators from surveys with satellite images and street view panorama.
However, These datasets rely on public health surveys, which limit their target scope. 
Recently, MedSat~\cite{scepanovic2024medsat} was developed in England by integrating sociodemographic features, satellite imagery, environmental variables, and prescription data. However, it is constrained by indirect prevalence estimation, leading to higher errors and limited disease scope. Finally, all these datasets suffered from a limited scope of target disease and none or moderate-resolution satellite images.
Inspired by all these works, we developed SatHealth by combining SDoH, environmental variables, satellite images, and the MarketScan medical claim database. High-resolution satellite images from Google Maps provide a clear view of ground-level environments, while MarketScan provides prevalence estimates for thousands of diseases. Hence, SatHealth enables comprehensive environment-health analyses.

\subsection{Health-related Target Modeling}

\textbf{Environment-health relationship modeling.} Numerous studies have been conducted to explore the impact of the environment on human health. The fundamental methods are community surveys and statistical tests. Wafula et al.~\cite{wafula_socioeconomic_2024} used surveys and mediation analysis to explore socioeconomic disparities in malaria prevalence in Malawi. Similarly, Keenan et al.~\cite{keenan_intersecting_2024} analyzed multidrug-resistant urinary tract infections (MDR UTIs) in East Africa using questionnaires and Bayesian profile regression to identify clusters of social and environmental determinants. By incorporating image modalities, Zhang et al.~\cite{zhang_higher_2024} explored the protective effects of shaded environments against adolescent myopia using commercial satellite maps and Spearman’s correlation analysis. 

Machine learning methods are also widely adopted. For example, Araújo et al.~\cite{araujo_earth_2024} employed Sentinel-1 and Sentinel-2 data to investigate the relationship between green spaces and mental health, using spatial autocorrelation metrics like Moran’s I and regression models to quantify the impact of environmental features. Yin et al.~\cite{yin_bayesian_2024} and Nazia et al.~\cite{nazia_identifying_2022} applied Bayesian hierarchical models to study the dynamics of COVID-19 spread, incorporating environmental, social, and health data. Luo et al.~\cite{luo_differing_2024} used mixed-effects models to analyze the association between air pollution and hypertension, revealing disparities in cardiovascular health risks. 
At the same time, Gibb et al.~\cite{gibb_interactions_2023} employed Bayesian spatiotemporal modeling to investigate dengue emergence linked to climate change and urban infrastructure. Most recently, Chen et al.~\cite{chen_artificial_2024} investigated the correlation between built environment from street view and coronary artery disease prevalence using deep learning-based features. They also explored the correlation between satellite imagery and cardiometabolic diseases~\cite{chen_deep_2024}. However, these works focus on some specific diseases without a broader understanding of the impact of the environment on diverse diseases.

\textbf{Personalized disease risk prediction.} Predictive risk modeling predicts patients' future disease status based on their historical medical records. Electronic health records (EHR) are generally used to provide patient medical history. As EHR data can be modeled naturally as sequential data, several deep-learning methods have been employed in previous studies for risk prediction. RETAIN~\cite{choi_retain_2016} uses reverse time attention to capture health status from the most recent patient visits. Dipole~\cite{ma_dipole_2017} uses a bidirectional recurrent neural network to capture more complex time dependencies in EHR. In recent years, there are also transformer-based models~\cite{yang_transformehr_2023, luo_hitanet_2020} and knowledge-enriched models~\cite{choi_gram_2017, ma_kame_2018}. However, these models focus solely on personal patient status without utilizing information from patients' living environments. We fill this gap by plugging our environmental embedding into patient representation according to their residence.

\section{SatHealth Dataset}

\label{sec:data_construction}


In this section, we introduce the SatHealth dataset by four components: SDoH, Environmental data, satellite imagery, and disease prevalence, as shown in \autoref{tab:var_stat}. Besides data collection, we will introduce our embedding pipeline for environmental data and satellite images.
We embed environmental data and satellite images across multiple geographic-level regions, including counties, ZIP Code Tabulation Areas (ZCTAs), census tracts, and Core Based Statistical Areas (CBSAs), ensuring comprehensive spatial granularity.

\begin{table}[t]
    \centering
    \caption{The number of variables in each category}
\vspace{-4mm}
    \label{tab:var_stat}
    \resizebox{0.47\textwidth}{!}{
    \begin{tabular}{ccll}
    \toprule
\multicolumn{1}{l}{}               & \multicolumn{1}{l}{Modality} & Category                               & Statistics            \\\midrule
SDoH                               & Tabular                      & SDI                                    & 8 Variables           \\\midrule
\multirow{4}{*}{Environmental}     & Tabular                      & Land Cover                             & 9 Variables           \\\cline{2-4}
                                   & \multirow{3}{*}{\makecell[c]{Time Series\\(Monthly)}} & Climate                                & 28 Variables, 7 years \\
                                   &                              & Air Quality                            & 4 Variables, 7 years  \\
                                   &                              & Greenery                               & 4 Variables, 7 years  \\\midrule
\multirow{3}{*}{Satellite Imagery} & \multirow{3}{*}{Image}       & \multicolumn{1}{c}{\multirow{3}{*}{-}} & 432918 Images         \\
                                   &                              & \multicolumn{1}{c}{}                   & 3 Channels (RGB)      \\
                                   &                              & \multicolumn{1}{c}{}                   & 9 Calculated Indices \\\midrule
\multirow{3}{*}{Medical Records}   & \multirow{3}{*}{\makecell[c]{Time Series\\(Yearly)}} & \multirow{3}{*}{\makecell[l]{ICD code\\Prevalence}}   & 1377 unique codes     \\
                                   &                              &                                        & 7 years coverage      \\
                                   &                              &                                        & 2141777 patients     \\\bottomrule
\end{tabular}}
\end{table}

\subsection{Social Determinants of Health}

Social determinants of health (SDoH) refer to the environmental conditions in which people live, influencing a wide range of health outcomes and quality of life~\cite{sdoh_def}. These include various socioeconomic factors, such as poverty, access to education, healthcare availability, the built environment, and community context. 

\textbf{Data Collection.} This work incorporates the Social Deprivation Index (SDI)~\cite{sdi_def} as the regional socioeconomic indicator. The SDI\footnote{\label{sdi_web}https://www.graham-center.org/maps-data-tools/social-deprivation-index.html} is a centile score comprised of seven demographic components derived from the American Community Survey (ACS): (1) the percentage of the population living below 100\% of the Federal Poverty Level (FPL); (2) the percentage of individuals aged 25 years or older with less than 12 years of education; (3) the percentage of non-employed individuals aged 16–64; (4) the percentage of single-parent families with dependents under 18; (5) the percentage of households without access to a vehicle; (6) the percentage of households residing in renter-occupied units; and (7) the percentage of households living in crowded conditions. 

\begin{table*}[t]
    \centering
    \caption{ICD codes with top Urban-rural relevance by Odds Ratio}
\vspace{-4mm}
    \label{tab:ur-test}
    \begin{tabular}{cllll}
    \toprule
\multicolumn{1}{l}{Category}      & Code    & Odds Ratio & 95\% CI        & Description                                                   \\\midrule
\multirow{2}{*}{Urban Prevalent}  & P00-P96 & 1.479     & (1.418, 1.542) & Certain conditions originating in the perinatal period        \\
                                  & O00-O9A & 1.160     & (1.131, 1.190) & Pregnancy, childbirth and the puerperium                      \\\midrule
\multirow{11}{*}{Rural Prevalent} & R99     & 0.312     & (0.227, 0.427) & Ill-defined and unknown cause of mortality                    \\\cline{2-5}
                                  & I00-I99 & 0.765     & (0.756, 0.773) & Diseases of the circulatory system                            \\
                                  & I05-I09 & 0.631     & (0.584, 0.681) & Chronic rheumatic heart diseases                              \\
                                  & I26-I28 & 0.735     & (0.687, 0.786) & Pulmonary heart disease and diseases of pulmonary circulation \\
                                  & I60-I69 & 0.736     & (0.701, 0.772) & Cerebrovascular diseases                                      \\
                                  & I10-I1A & 0.743     & (0.735, 0.752) & Hypertensive diseases                                         \\\cline{2-5}
                                  & E00-E89 & 0.785     & (0.777, 0.793) & Endocrine, nutritional and metabolic diseases                 \\
                                  & E40-E46 & 0.636     & (0.583, 0.694) & Malnutrition                                                  \\
                                  & E70-E88 & 0.737     & (0.728, 0.745) & Metabolic disorders                                           \\
                                  & E65-E68 & 0.773     & (0.761, 0.785) & Overweight, obesity and other hyperalimentation               \\
                                  & E08-E13 & 0.792     & (0.778, 0.806) & Diabetes mellitus  \\\bottomrule
\end{tabular}
\end{table*}

\begin{figure*}[ht]
    \vspace{-0.25cm}
    \centering
    \includegraphics[width=0.95\linewidth]{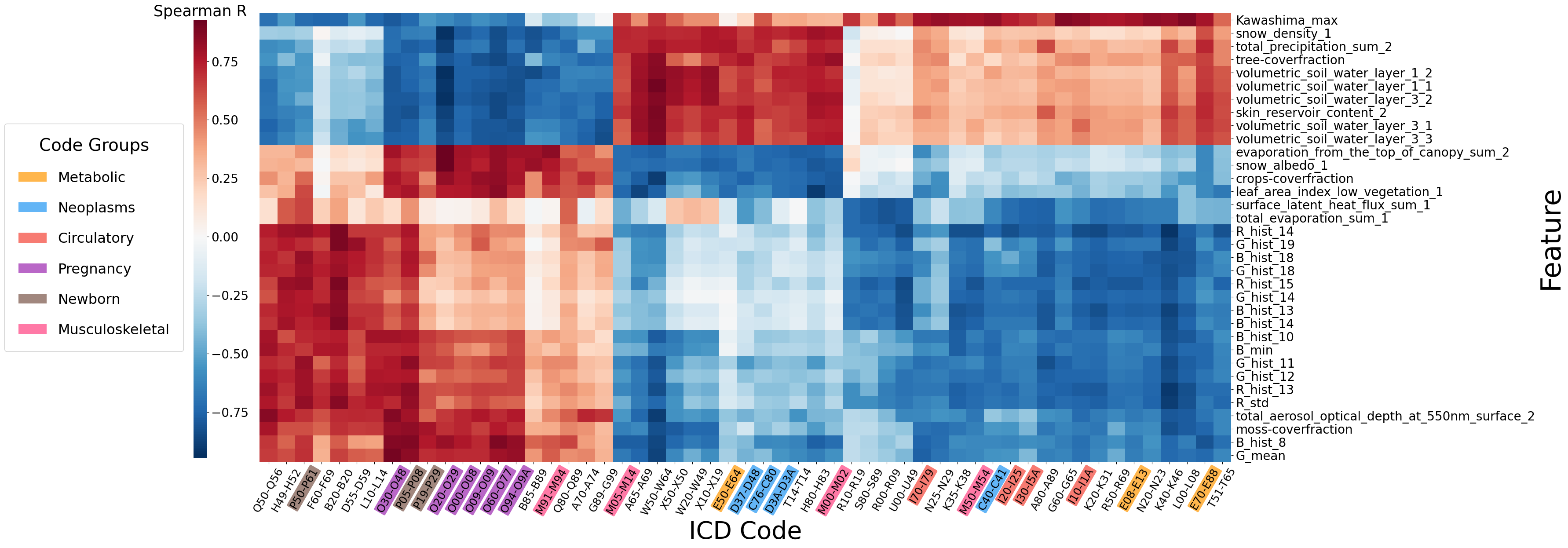}
    \vspace{-0.35cm}
    \caption{Feature correlations - second level ICD}
    \label{fig:feat-corr2}
\end{figure*}

\subsection{Environmental Data} Environmental data contains four categories: climate, air quality, greenery, and land cover. We collect them from multiple satellite products.

\textbf{Data Collection.} We obtained environmental variables for our dataset on Google Earth Engine (GEE)~\cite{gorelick_google_2017} following~\cite{scepanovic2024medsat}. Climate variables include temperatures, humidity, solar radiation, snow cover, and wind components from ERA5-ECMWF product~\cite{era5_data}. For air quality, we collect nitrogen dioxide (NO2) from Sentinel-5P Near Real-Time NO2~\cite{irizar2019sentinel}, total aerosols and PM2.5 from Copernicus Atmosphere Monitoring Service (CAMS)~\cite{garrigues2022monitoring}, and ozone from Total Ozone Mapping Spectrometer (TOMS) and Ozone Monitoring Instrument (OMI) Merged Ozone Data~\cite{ahmad2006aerosol}.  Greenary variables include Normalized Difference Vegetation Index (NDVI) derived from Sentinel-2 MultiSpectral Instrument (MSI)~\cite{sentinel_2_info} and high/low vegetation greenery indices from ERA5-ECMWF. Land cover variables are area cover fractions of different land types (\eg forest, water, urban) sourced from Copernicus Dynamic Land Cover products~\cite{copernicus_dynamic_landcover}. We collected all the available data in Ohio from 2016 to 2022.

\textbf{Spatiotemporal Alignment.} As shown in Table \ref{tab:supp_data_license}, data from various satellite products have different spatial resolutions and temporal frequencies, so we align them to ensure consistency in both spatial and temporal granularity.
Given a specific timestamp and an environment variable, we collect the values in various areas and obtain a heatmap, with each pixel denoting the value in a square area. We perform spatial reduction to a specific region (\eg counties) by taking the average of all pixels within the region. After the reduction for every variable, we obtain an environmental vector for each specific region. We further align variables temporally by downsampling all vector series to the same monthly frequency. As land cover fractions are relatively stable, they are averaged across all timestamps to become a static variable.

\textbf{Feature Embedding.} After the alignment process, each region has environment data as a multivariate time series. To embed the time series, we average the monthly vectors according to meteorological seasons and concatenate seasonal vectors to create annual regional embedding. 


\subsection{Satellite Images}

In addition to environmental variables, satellite images offer additional visual insights into specific regions. By incorporating satellite imagery from Google Maps, we provide an implicit indicator of regional development, environmental characteristics, and land-use patterns. 

\textbf{Data Collection.} We request aerial images from the Google Static Maps Application Programming Interface (API)~\cite{google_static_map}. A grid of spatial points with 500-meter spacing is created, and satellite image patches are retrieved at zoom level 17 with grid points. With the point-grid-based construction, we can establish the visual feature of an arbitrary region by aggregating satellite images within that region. This approach ensures complete coverage of Ohio, resulting in 432,918 images with a resolution of 1280$\times$1280 pixels each, corresponding to a square area approximately 500 meters wide.

\textbf{Feature Embedding.} We establish visual features by calculating several indices commonly used in remote sensing~\cite{zheng_evaluation_2018, gee_rgb_2020, cao_comparison_2021, feng_comparison_2022} and computing their pixel-level statistics. For a given RGB aerial image, nine indices are calculated for each pixel based on RGB values, effectively creating new derived channels. These channels form a compound image with 12 channels alongside the original RGB channels. After that, to extract meaningful features from the compound images, we compute pixel-level statistics for each channel, including the mean, standard deviation, median, maximum, minimum, and 20-bin histogram features. This process produces feature vectors of size 300 for each satellite image. Note that images from Google Maps do not have timestamps, so the satellite image feature is static.

\subsection{Disease Prevalence} We use regional disease prevalence as an indicator of the population health status of an area. 
  
\textbf{Data Collection.} We estimate disease prevalence using MarketScan~\cite{merative_real_2024}, a real-world medical claims database. Specifically, we analyze patient encounters from the MarketScan Commercial Claims and Encounters (CCAE) database from 2016 to 2022. Each patient encounter includes a set of International Classification of Diseases (ICD-10)~\cite{cdc_icd-10-cm_2024} diagnoses codes and the patient's residency. Patient residency in MarketScan is identified by Metropolitan Statistical Areas (MSAs), a subset of CBSAs. This enables us to estimate the prevalence of each ICD-10 code for MSAs in a given year by calculating the percentage of patients associated with the code among all patients recorded in that year.

\textbf{Data Processing.} The ICD-10 codes are organized hierarchically, allowing for a multi-level investigation of disease prevalence. For instance, the code "I10" represents essential hypertension, its parent node "I10-I1A" represents a broader group of hypertensive diseases, and the top-level code "I00-I99" encompasses circulatory system diseases. This hierarchical structure allows us to explore the prevalence of both broad disease categories and specific conditions. To capture both the overall trends and finer details, we calculated disease prevalence at the top three levels of ICD-10 codes using their first three digits, enabling a comprehensive understanding of disease patterns and correlations.

\section{Dataset Use Cases}



In this section, we first conduct a comprehensive analysis to examine the correlation between environmental data and human health status. We then demonstrate potential use cases of SatHealth through two kinds of tasks: regional public health modeling and personalized disease risk prediction. For regional public health modeling, we use regional environmental embeddings to predict the SDI score and disease prevalences, which display the power of multimodal data in modeling regional health status. For personalized disease risk prediction, we plug the environmental factors into patient representations produced by EHR risk prediction backbones according to patient residency, which shows the benefits of environmental information in predicting personal disease risks. 

\begin{table*}[ht]
\centering
\caption{Regression performance for SDI components and disease prevalence}
\vspace{-4mm}
\label{tab:pred_region}
\resizebox{\textwidth}{!}{
    \begin{tabular}{c|l|llll|llll|llll}
    \toprule
    &                         & \multicolumn{4}{c|}{MAE$\;\downarrow$}                                                                                                & \multicolumn{4}{c|}{MSE$\;\downarrow$}                                                                                                & \multicolumn{4}{c}{R$^2\;\uparrow$}                                                                                                 \\\midrule
    Category&       Target          & \multicolumn{1}{c}{DEnv} & \multicolumn{1}{c}{LC} & \multicolumn{1}{c}{Img} & \multicolumn{1}{c|}{All} & \multicolumn{1}{c}{DEnv} & \multicolumn{1}{c}{LC} & \multicolumn{1}{c}{Img} & \multicolumn{1}{c|}{All} & \multicolumn{1}{c}{DEnv} & \multicolumn{1}{c}{LC} & \multicolumn{1}{c}{Img} & \multicolumn{1}{c}{All} \\\midrule
    \multirow{8}{*}{SDoH}& SDI                       & \textbf{0.564}              & 0.621                         & 0.618                     & \textbf{0.564}                  & \textbf{0.461}              & 0.590                         & 0.544                     & 0.464                        & \textbf{0.558}              & 0.434                         & 0.478                     & 0.555                        \\
    &-$\;$pct$\_$Poverty$\_$LT100       & 0.516                       & \textbf{0.473}                   & 0.480                & \textbf{0.473}                 & 0.567                       & 0.424                         & \textbf{0.403}            & 0.413                        & 0.277                       & 0.458                         & \textbf{0.486}            & 0.473                        \\
    &-$\;$pct$\_$Education$\_$LT12years & \textbf{0.495}              & 0.600                         & 0.572                     & 0.536                        & \textbf{0.526}              & 0.762                         & 0.606                     & 0.552                        & \textbf{0.343}              & 0.048                         & 0.243                     & 0.311                        \\
    &-$\;$pct$\_$NonEmployed          & 0.533                       & 0.593                         & 0.560                     & \textbf{0.532}               & 0.740                       & 0.679                         & 0.659                     & \textbf{0.577}               & 0.140                       & 0.210                         & 0.234                     & \textbf{0.329}               \\
    &-$\;$pct$\_$Single$\_$Parent$\_$Fam  & 0.509                       & 0.443                         & \textbf{0.414}            & 0.433                        & 0.680                       & 0.424                         & \textbf{0.369}            & 0.396                        & -0.285                      & 0.199                         & \textbf{0.302}            & 0.251                        \\
    &-$\;$pctHH$\_$No$\_$Vehicle        & 0.484                       & 0.512                         & 0.519                     & \textbf{0.438}               & 0.709                       & 0.743                         & 0.675                     & \textbf{0.562}               & 0.406                       & 0.377                         & 0.434                     & \textbf{0.529}               \\
    &-$\;$pctHH$\_$Renter$\_$Occupied   & 0.445                       & 0.462                         & 0.447                     & \textbf{0.406}               & 0.382                       & 0.444                         & 0.366                     & \textbf{0.315}               & 0.580                       & 0.513                         & 0.598                     & \textbf{0.654}               \\
    &-$\;$pctHH$\_$Crowding           & 0.682                       & 0.709                         & 0.697                     & \textbf{0.675}               & 1.113                       & 1.222                         & 1.192                     & \textbf{1.052}               & 0.009                       & -0.089                        & -0.063                    & \textbf{0.062}    \\\midrule
    \multirow{7}{*}{\makecell[c]{Disease\\Prevalence}}& Neoplasms                   & 0.389 & 0.392          & 0.391          & \textbf{0.327} & 0.284 & 0.288 & 0.284          & \textbf{0.215} & 0.657 & 0.621 & 0.632          & \textbf{0.743} \\
& Metabolic diseases          & 0.424 & 0.384          & 0.385          & \textbf{0.366} & 0.299 & 0.248 & 0.250          & \textbf{0.228} & 0.686 & 0.706 & 0.704          & \textbf{0.742} \\
& -$\;$Diabetes mellitus           & 0.295 & 0.268          & 0.271          & \textbf{0.267} & 0.158 & 0.137 & 0.139          & \textbf{0.127} & 0.831 & 0.848 & 0.845          & \textbf{0.861} \\
& -$\;$Metabolic disorders         & 0.326 & \textbf{0.249} & 0.250          & 0.284          & 0.195 & 0.116 & \textbf{0.115} & 0.142          & 0.781 & 0.845 & \textbf{0.846} & 0.827          \\
& Circulatory system diseases & 0.227 & 0.205          & \textbf{0.200} & 0.212          & 0.086 & 0.075 & \textbf{0.072} & 0.078          & 0.900 & 0.909 & \textbf{0.912} & 0.909          \\
& -$\;$Hypertensive diseases       & 0.198 & 0.194          & 0.186          & \textbf{0.177} & 0.063 & 0.062 & 0.058          & \textbf{0.052} & 0.928 & 0.927 & 0.930          & \textbf{0.939} \\
& -$\;$Ischemic heart diseases     & 0.304 & \textbf{0.249} & 0.250          & 0.254          & 0.185 & 0.123 & \textbf{0.122} & 0.142          & 0.815 & 0.873 & \textbf{0.874} & 0.855         \\\bottomrule
    \multicolumn{13}{l}{All: embeddings by combining all modalities; "-" denotes subcategories}\\
    \end{tabular}
}
\end{table*}

\subsection{Basic Correlation Analysis}
We first perform a statistical analysis to analyze the correlation between the living environment and regional disease prevalence. Note that this is a simple illustration of the dataset characteristics instead of a rigorous public health analysis.

\subsubsection{Regional Disparity}

We start by exploring the disparity in health status between regions in Ohio. For a simple illustration, we define Columbus, Cleveland, and Cincinnati as urban areas and compare the disease occurrence with other areas in Ohio. Specifically, we calculate urban-to-rural odds ratios (OR) and their confidence intervals (CI) to show the difference in disease occurrence between the urban (Columbus, Cleveland, and Cincinnati) and other areas. 

As \autoref{tab:ur-test} shows, the urban areas present a significantly higher prevalence of neonatal conditions and pregnancy-related codes. Specifically, the conditions originating in the perinatal period (P00-P96) have an odds ratio of 1.479 (95\% CI: 1.418-1.542), and codes for pregnancy, childbirth, and the puerperium (O00-O9A) are also more frequent in urban areas (OR: 1.16, 95\% CI: 1.131-1.190). This disparity could have originated from the higher in-hospital ratio of births~\cite{way_out-hospital_2022} and limited access to maternity and prenatal care~\cite{yin_urbanrural_2019, wendling_access_2021, dipietro_mager_routine_2021}. 

Moreover, rural areas have more mortality cases with ill-defined and unknown causes (ICD-10: R99, OR: 0.311, 95\% CI: 0.227-0.427). This kind of coding is less informative and reflects a relatively lower data quality from the rural healthcare system~\cite{mikkelsen_assessing_2020, who_ill-defined_2025}. Besides, as the US Centers for Disease Control and Prevention (CDC) reported~\cite{cdc_about_2024}, rural residents tend to be older and sicker, and they have higher rates of cigarette smoking, high blood pressure, and obesity. Correspondingly, we observed higher prevalences of circulatory system diseases (I00-I99) as well as Endocrine, nutritional, and metabolic diseases (E00-E89). Specifically, rural areas have higher prevalences for chronic rheumatic heart diseases (OR: 0.631, 95\% CI: 0.584-0.681), hypertensive diseases (OR: 0.743, 95\% CI: 0.735-0.752), obesity (OR: 0.773, 95\% CI: 0.761-0.785), and diabetes (OR: 0.792, 95\% CI: 0.778-0.806). We show these results in Tabel \ref{tab:ur-test}, and more comprehensive results can be found in Table \ref{tab:supp_ur-test} in supplementary. These findings help us in discovering health inequalities and support policy making~\cite{loccoh_rural-urban_2022, yeh_sustainbench_2021}. Furthermore, such observations and findings from public health data can help policy development, such as encouraging network development and telemedicine, and improving the rules for Medicare payments to providers~\cite{ricketts_changing_2000, coughlin_continuing_2019}.

\subsubsection{Factor Correlations}

In addition to the odds ratio, we calculate Spearman's rank correlation coefficient to investigate the correlation between environmental factors and multi-level ICD code prevalence across spatial regions. We show the pairwise correlations between level-2 ICD codes and environmental features in Figure \ref{fig:feat-corr2}. Note that this figure only shows subsets of ICD codes and features with significant correlations. Additionally, we use color shading on ICD codes to differentiate between disease categories. More detailed results on a broader set of ICD codes are shown in Figure \ref{fig:feat-corr} and \ref{fig:feat-corr1} in the supplementary results.

We observed several outstanding color blocks in Figure \ref{fig:feat-corr2}, showing groups of diseases with distinct correlations with specific feature subsets. Firstly, a group of newborn or pregnancy-related conditions (O and P codes) with a purple or brown background lies on the left-hand side, with a significant negative correlation with tree cover fraction and volumetric soil water. Such correlation can be a reflection of the higher prevalence of newborn or pregnancy-related conditions in urban areas we found previously, as urban soil tends to be compact with lower water content~\cite{urban_soil}. In contrast, several neoplasm conditions show the opposite correlation pattern, which reflects a higher prevalence of tumors in rural areas~\cite{semprini_ruralurban_2024}. Moreover, the correlation between neoplasms and soil water may also be explained by the higher mobility of chemical pollutants with higher precipitation and moisturized soil~\cite{biswas_fate_2018}. 

Moreover, some circulatory system diseases in orange color on the right-hand side, such as hypertensive diseases (I10-I1A) and heart diseases (I30-I5A). A similar pattern also applies to diabetes mellitus (E08-E13) and metabolic disorders (E70-E88). These diseases correlate positively with the Kawashima index~\cite{kawashima_algorithm_1998} and negatively with many satellite image features. As the Kawashima index is negatively correlated to chlorophyll content~\cite{signorelli_green_2023}, and image features also provide information about pixel color distribution, these correlations could be explained by the discovered benefit of green space on cardiovascular disease and diabetes~\cite{astell-burt_urban_2019, asri_global_2020}. 
\begin{table*}
\centering
\caption{Prediction performance of personalized disease risk prediction}
\vspace{-4mm}
\label{tab:person-pred}
\resizebox{\linewidth}{!}{
    \begin{tabular}{ll|ccccc|ccccc}
    \toprule
                                         &      & \multicolumn{5}{c|}{Next Visit Prediction}                        & \multicolumn{5}{c}{1-Year Diagnosis Code Prediction}                 \\
                                         &      & mAUC$\uparrow$ & mAUC-t10$\uparrow$ & Recall@5$\uparrow$ & Recall@10$\uparrow$ & Recall@50$\uparrow$ & mAUC$\uparrow$ & mAUC-t10$\uparrow$ & Recall@5$\uparrow$ & Recall@10$\uparrow$ & Recall@50$\uparrow$ \\\midrule
\multirow{2}{*}{LSTM}        & EHR  & 0.481          & 0.745          & 0.240          & 0.253          & 0.416          & 0.487          & 0.770          & \textbf{0.503} & \textbf{0.537} & \textbf{0.687} \\
                             & +Env & \textbf{0.538} & \textbf{0.767} & \textbf{0.288} & \textbf{0.325} & \textbf{0.561} & \textbf{0.551} & \textbf{0.815} & 0.489          & 0.508          & 0.642          \\\midrule
\multirow{2}{*}{RETAIN}      & EHR  & 0.476          & 0.678          & \textbf{0.273}          & \textbf{0.301} & \textbf{0.456} & 0.492          & 0.703          & 0.198          & 0.221          & 0.463          \\
                             & +Env & \textbf{0.551} & \textbf{0.816} & \textbf{0.273} & 0.287          & 0.445          & \textbf{0.543} & \textbf{0.764} & \textbf{0.205} & \textbf{0.265} & \textbf{0.503} \\\midrule
\multirow{2}{*}{Dipole}      & EHR  & 0.600          & 0.869          & 0.263          & 0.302          & 0.487          & 0.660          & 0.929          & 0.477          & 0.504          & 0.641          \\
                             & +Env & \textbf{0.722} & \textbf{0.942} & \textbf{0.416} & \textbf{0.480} & \textbf{0.724} & \textbf{0.734} & \textbf{0.932} & \textbf{0.553} & \textbf{0.587} & \textbf{0.769} \\\midrule
\multirow{2}{*}{Transformer} & EHR  & 0.484          & 0.679          & 0.215          & 0.240          & 0.439          & 0.503          & 0.706          & 0.310          & 0.333          & 0.523          \\
                             & +Env & \textbf{0.549} & \textbf{0.738} & \textbf{0.234} & \textbf{0.265} & \textbf{0.495} & \textbf{0.553} & \textbf{0.732} & \textbf{0.341} & \textbf{0.377} & \textbf{0.584}   \\\bottomrule      
\end{tabular}}
\end{table*}

These results highlight the significant impact of the living environment on human health, consistent with existing studies~\cite{keenan_intersecting_2024, wafula_socioeconomic_2024, spoerri_swiss_2010}. They also demonstrate the potential of SatHealth to enhance AI models for health-related tasks (\eg regional public health modeling and personalized disease risk prediction).

\subsection{Regional Public Health Modeling}
\label{sec:regional_public_health_modeling}

Regional public health modeling aims to predict health outcomes for the community or regional population. It helps identify health disparities and guide targeted interventions to improve public health outcomes and resource allocation~\cite{loccoh_rural-urban_2022, yeh_sustainbench_2021}. In this subsection, we use the created regional environment embeddings to predict regional SDI and disease prevalence to reveal the relationship between living environment and population health status.


We divide the environmental features into two subsets according to data structure: (1) \textbf{Dynamic environmental factors (DEnv)} refers to environmental factors that change over time, including climate, air quality, and greenery variables; (2) \textbf{Static environmental factors} refers to environmental factors that are stable over time, which include \textbf{Land Cover (LC)} and \textbf{Satellite images (Img)}. 
For baselines, we train random forest regressors respectively on the feature subsets to show their fundamental functional relevance. After that, we combine all modalities \textbf{(All)} and display how this helps in regional health status modeling.


\subsubsection{SDoH Regression} The upper part of \autoref{tab:pred_region} presents the performance of the Social Deprivation Index (SDI) prediction. The performance metrics include Mean Absolute Error (MAE), Mean Squared Error (MSE), and $R^2$, providing a comprehensive assessment of model accuracy and explanatory power. It is worth pointing out that we normalize the prevalence target by z-score normalization such that they have a standardized deviation of 1.

\textbf{Overall SDI.} When predicting overall SDI, the dynamic features (DEnv) achieve the best performance, yielding the lowest MAE and highest $R^2$. Besides, the combined feature (All) performs slightly worse but is comparable to DEnv, suggesting that DEnv variables predominantly drive the prediction. In contrast, land cover and image embeddings show suboptimal performance, likely because they capture only ground-level views, making it challenging to predict a comprehensive socioeconomic score.  


\textbf{Population Characteristics} include poverty, education, and employment status. In this category, the best-performing features vary, suggesting different dependencies of SDoH on environmental factors. Specifically, the poverty ratio is better predicted by image features, likely due to their ability to capture urban-rural disparities, such as housing conditions. Education status relies more on dynamic features (\ie climate, greenery, air quality). Employment benefits most from multimodal fusion as a combination of economic, infrastructure, and environmental factors influences it.


\textbf{Household Characteristics} refer to household-level factors, including single-parent family percentage, housing tenure (rent or owner-occupied), vehicle ownership, and household crowdedness. The combined feature outperforms all other features in predicting household characteristics except for the single-parent family ratio. This superiority indicates that integrating multimodal features provides a more comprehensive understanding of housing-related metrics. The combined features perform poorly in predicting the single-parent family ratio, likely because dynamic features (DEnv) tend to be independent of family structure. However, it is worth noting that none of these features are effective predictors for crowding prediction, with the best $R^2$ score around 0.06. 

 
\subsubsection{Disease Prevalence Regression}

There are thousands of diseases, but only a small subset correlates significantly with environmental features. Therefore, we focus on diseases reported to correlate with environmental factors~\cite{astell-burt_urban_2019, asri_global_2020}: neoplasms (C00-D49), endocrine, nutritional, and metabolic diseases (E00-E89), and circulatory system diseases (I00-I99). Moreover, we explored several subcategories, including diabetes (E08-E13), metabolic disorders (E70-E88), hypertensive diseases (I10-I1A), and ischemic heart diseases (I20-I25). 

\textbf{Overall Comparison.} We show the regression performance of predicting the prevalence of the diseases in the lower part of \autoref{tab:pred_region}. 
For most diseases (\eg Neoplasms and Metabolic diseases), combining all modalities (All) achieves the best prediction performance, demonstrating the essential role of multimodal environmental factors in modeling the regional prevalence of these diseases.

\textbf{Neoplasms}. Neoplasms (C00-D49) include all kinds of tumors, including both malignant and benign.
\autoref{tab:pred_region} shows that dynamic features (DEnv) have better prediction performance than land cover and satellite, as these features contain carcinogens potentially related to some specific cancer, such as solar radiation~\cite{asri_global_2020} and air pollutants~\cite{wang_association_2022}. Moreover, by combining all features, the regression model gains 0.086 improvement in $R^2$, the highest improvement among the explored diseases. Such improvement demonstrates that the three feature sets are complementary when modeling neoplasm prevalence.

\textbf{Metabolic diseases.} As for endocrine, nutritional, and metabolic diseases (E00-E89), land cover (LC) and satellite image (Img) features perform consistently better than dynamic factors (DEnv). Especially for metabolic disorders (E70-E88), the $R^2$ score on dynamic factors is 0.064 lower than the other two features. This is intuitively reasonable as these diseases are more related to our lifestyle~\cite{lam_built_2023, patwary_impact_2024}, and the effect of air quality or climate tends to be long-term and maybe implicit~\cite{zheng_effects_2022}. 

\textbf{Circulatory system diseases.} Finally, it can be seen that the three feature modalities gain high but similar performance for circulatory system diseases (I00-I99), and the improvement of combined features is not significant. This could be due to each feature modality's relatively high performance, making it hard to improve further. On the other hand, land cover (LC) and satellite features (Img) have higher performances than dynamic factors (DEnv) for ischemic heart diseases (I20-I25), possibly due to the reported correlation between green space and ischemic heart diseases~\cite{astell-burt_urban_2019}.



\begin{table}[t]
\centering
\caption{Model $R^2$ scores for Spatiotemporal Generalization.  Bold: best results, \underline{underlined}: second best.}
\label{tab:ts-valid}
    \begin{tabular}{lcc|cc|c}
    \toprule
             & \multicolumn{2}{c|}{Spa. In.} & \multicolumn{2}{c|}{Spa. Ex.} & \multicolumn{1}{c}{Temp. For.} \\\midrule
Features & \multicolumn{1}{c}{SDoH} & \multicolumn{1}{c|}{Prev.} & \multicolumn{1}{c}{SDoH} & \multicolumn{1}{c|}{Prev.} & \multicolumn{1}{c}{Prev.}     \\\midrule
DEnv         & 0.183                    & 0.667                          & -0.186                   & 0.552                          & 0.811                              \\
DEnv + T     & 0.161                    & 0.665                          & -0.301                   & 0.522                          & 0.868                              \\
DEnv + S     & 0.183                    & 0.705                          & -0.261                   & 0.552                          & 0.759                              \\
DEnv + T + S & 0.165                    & \textbf{0.711}                 & -0.296                   & \textbf{0.624}                 & 0.820                              \\\midrule
All          & 0.244                    & 0.665                          & \textbf{0.035}           & 0.543                          & 0.902                              \\
All + T      & 0.238                    & 0.671                          & -0.005                   & 0.525                          & \textbf{0.928}                     \\
All + S      & \textbf{0.275}           & 0.699                          & 0.014                    & 0.572                          & 0.879                              \\
All + T + S  & {\ul 0.265}              & {\ul 0.709}                    & {\ul 0.026}              & {\ul 0.619}                    & {\ul 0.914}                        \\\bottomrule
\multicolumn{5}{l}{\small\makecell[l]{Prev.: Disease Prevalence; Spa.: Spatial; Temp.: Temporal;\\Ex.: extrapolation; In.: interpolation; For.: forecasting}}  \\      
\end{tabular}
\end{table}

\subsection{Personalized Disease Risk Prediction}
In this subsection, we use living environmental data to help predict patient-level disease risks. Specifically, we focus on the next visit prediction and 1-year predictive modeling tasks.
We use the patient visit history data from the MarketScan database~\cite{chen_predictive_2024} to conduct experiments.
We build models to predict the level-2 ICD codes within a patient visit with all previous visits from the patient. We define the multi-label ground truth in two ways: \textbf{Next visit} setting takes the codes from one patient as ground truth no matter how long it is from the patient's last visit, while \textbf{1-Year Predictive Modeling} setting takes all code within the next one year of the patient's last known visit. In addition to patients' visit history, we concatenate the embedding of the patient's living environment to the patient representation produced by the visit encoder network to enhance the backbone model. The embedding of the patient's living environment is created by data within the patient's residence area in the year of the patient's last known visit. 

We implemented several backbone networks for comparison. They are well-known architectures handling sequences, including LSTM~\cite{hochreiter_long_1997} and Transformer~\cite{vaswani_attention_2017}, and models designed for medical data such as RETAIN~\cite{choi_retain_2016} and Dipole~\cite{ma_dipole_2017}. This task can be formulated as a multi-label problem, so we use the macro average AUROC score (mAUC) across all diseases as the basic evaluation metric. Additionally, we use the macro average AUROC score of the top 10 performed diseases (mAUC-t10). Furthermore, as this model predicts the disease risk for each patient, we calculate the recall at k (Recall@k) metric to evaluate the model's capability of producing correct disease warnings for patients. 

We put the experimental results in Table \ref{tab:person-pred}. We highlight the best performance among two variations in bold for each backbone model. 
Using environmental information boosts model performance in most cases, especially for the recall, indicating that the patient's living environment may help us identify their disease risks. Moreover, for the next visit prediction task, models with environmental information tend to have much higher mAUC scores for top-performed diseases, which shows the capability of environmental data in differentiating patients with various disease risks.

\subsection{Spatiotemporal Generalizability Analysis}

\begin{figure}[t]
    \centering
    \includegraphics[width=0.98\linewidth]{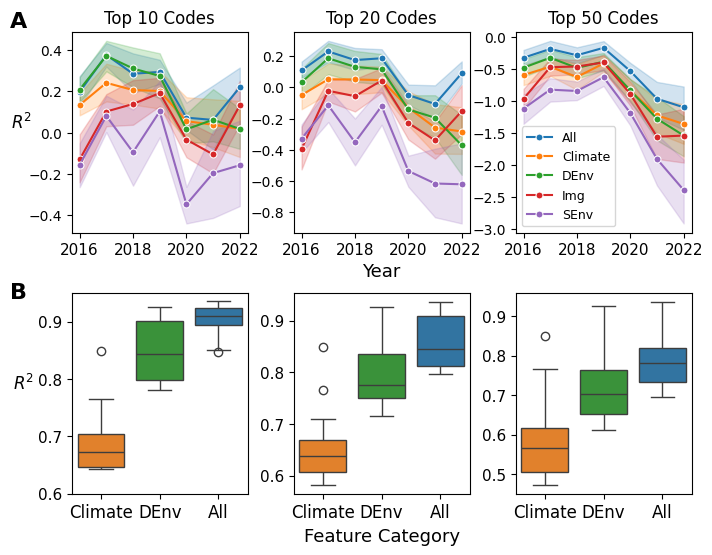}
    \caption{Random Forest Regression Performance from Different Feature Groups. (A) $R^2$ score per year for spatial interpolation. (B) $R^2$ score for temporal extrapolation. }
    \label{fig:perf-icd}
\end{figure}

The living environment varies in different areas and is subject to change over time. Therefore, users must be careful about spatiotemporal distribution shifts when applying models trained on SatHealth to other regions or conducting long-term forecasting. To find out how this affects the model performance with SatHealth, we conducted experiments on three cases of spatiotemporal generalization on regional public health modeling. Moreover, we designed a spatiotemporal-enhanced regression strategy to improve model robustness. In this subsection, we show the experimental performance of the enhanced model in the three generalization scenarios: spatial interpolation (\eg missing value imputation~\cite{baker_missing_2014}), spatial extrapolation (\eg distant region generalization~\cite{patton_transferability_2015, meyer_predicting_2021}), and temporal forecasting~\cite{nikparvar_spatio-temporal_2021, heuton_spatiotemporal_2024}. We provide more experimental details in the supplementary Section \ref{sec:ts_gen}.

We start by evaluating the generalizability of basic regression models without spatiotemporal enhancements. Using $R^2$ scores, we show the top 10, 20, and 50 performing diseases. Figure \ref{fig:perf-icd} compares model performance across input modalities for (A) spatial interpolation and (B) temporal forecasting. LC and Img models are excluded from temporal forecasting due to their time independence. A model trained solely on climate variables is also included for comparison. For spatial interpolation, we stratify results by year. As shown in Figure \ref{fig:perf-icd}B, Combined features show higher stability to temporal shift, likely due to reduced feature variance over time. Additionally, the combined features improve spatial generalization. Dynamic features, including climate, greenery, and air quality, exhibit stronger spatial robustness, possibly due to their pronounced spatial clustering effect.

Next, we show how spatiotemporal information can improve model generalizability. We train boosting models as described in Section \ref{sec:ts_regression}. Models with neighborhood information are annotated with "+S", and models with history information are annotated with "+T". We compared the combined feature model (All) with the Dynamic feature model (DEnv) on both SDoH and disease prevalence regression under three generalization cases, and the result is shown in Table \ref{tab:ts-valid}. It can be seen that incorporating spatiotemporal information can be beneficial to model robustness, as the "+T+S" models are either the best or the second best for all cases. Although not the best, the "All+T+S" model is comparable to the best, providing a general solution that works under all cases. Moreover, the combined feature models tend to perform better than the DEnv models, further showing the multimodal models' benefits. Finally, we find that our model performs poorly when predicting SDoH under spatial extrapolation, which may indicate a high spatial disparity between regions, making the spatial generalization of the SDoH model challenging. 
\section{Web Application Deployment}

\begin{figure}[t]
    \centering
    \includegraphics[width=\linewidth]{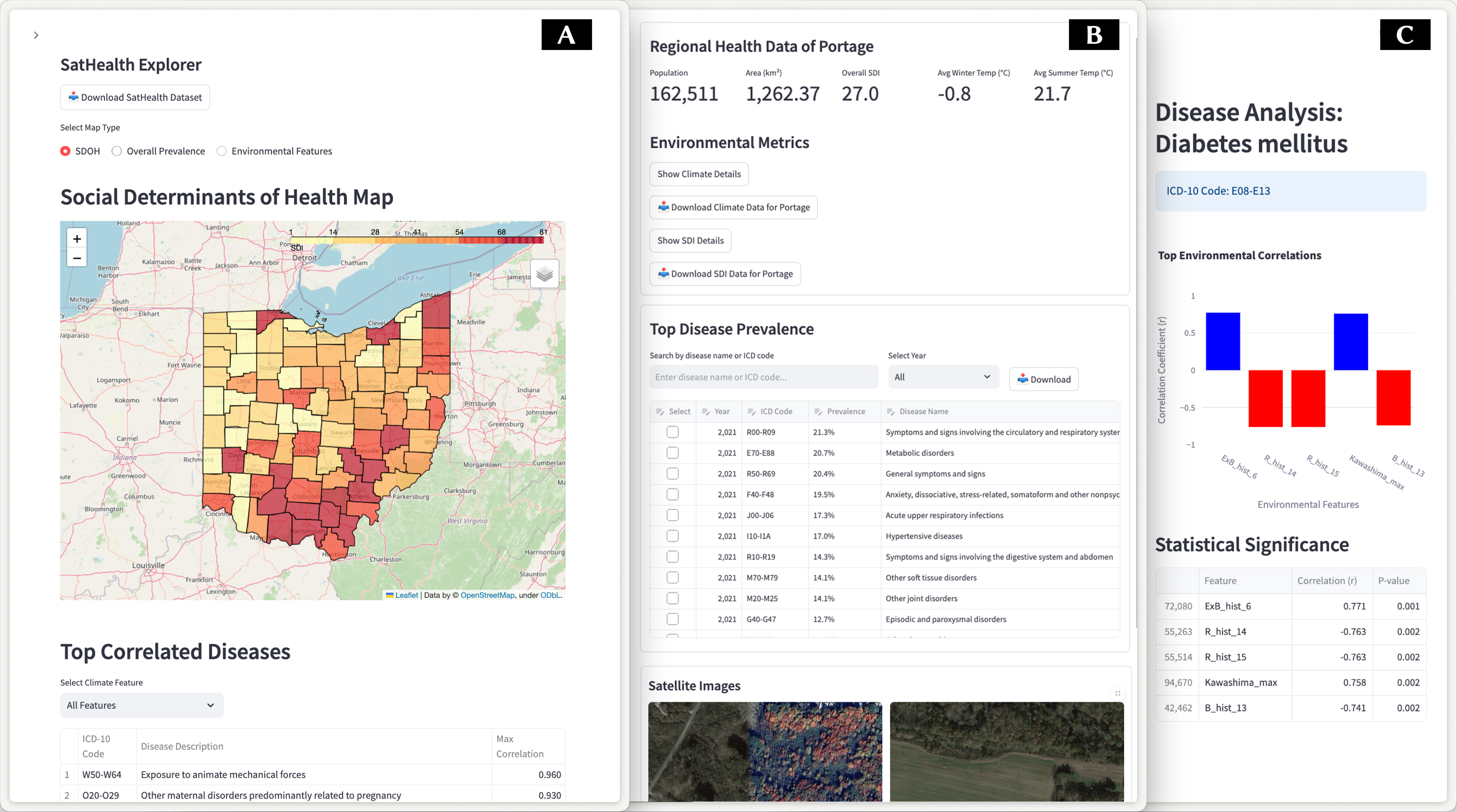}
\vspace{-3mm}
    \caption{UI design of the proposed dataset}
\vspace{-2mm}
    \label{fig:ui_design}
\end{figure}

To improve the utility of our dataset, we designed a web-based application to explore and access SatHealth. The system design is shown in Figure \ref{fig:ui_design}.

The default page (Figure \ref{fig:ui_design}A) is the satellite map of Ohio, with the overall SDI score for each county shown in the heatmap. Users can select different variables to be displayed in the heatmap, including SDI, environmental variables, and disease prevalence of the user-selected ICD code. For time-dependent variables, there's a slide bar for users to define the time range of the data shown. Moreover, diseases with a high correlation to environmental features are shown below the map; users can also select specific environmental features to check their correlation to the map. We also provide the button to download the whole dataset at the top of this page. More information about data access can be found in Section \ref{sec:supp_data_access}.

Once users click on one of the regions in the map, its basic information will be shown in a new column on the right-hand side (Figure \ref{fig:ui_design}B). There will be three sub-panels in this sidebar. On the top, basic metrics, including population and total area of the region, are shown, together with the overall SDI score and seasonal average temperature. When users click the "Show Details" button, a drop-down list will appear to display more variables. Buttons are also provided to download the SatHealth subset in the selected region. In the middle, there's a list of the most prevalent diseases in the selected region; users can also search for ICD codes to see their prevalence. At the bottom of this sidebar, random satellite images in the selected region will be displayed.

Users can check the selection box in the disease list. The right-hand side column will be updated (Figure \ref{fig:ui_design}C) to show a bar chart of the correlation score of the selected disease with the top 5 correlated environmental features in a bar chart. The correlation score and corresponding p-value will also be displayed below the bar chart. Users can search for a specific variable to check its correlation to the current disease. 

\section{Limitations and Future Work}

Our work still has several limitations. First, SatHealth is currently restricted to Ohio due to budget constraints, but we will gradually expand our dataset to include more states, toward full US coverage. Furthermore, although environmental data showed effectiveness in our experiment, our embedding method of such multimodal data is relatively straightforward using feature engineering. We will explore more complicated deep-learning strategies to unlock the power of multimodal environmental data. Finally, the individual residences in our dataset are coarse due to privacy issues, making the living environment less representative. Despite these limitations, we believe SatHealth still provides a solid foundation and a good starting point for research on the interplay between the living environment and human health.

\section{Conclusion} 
In this study, we developed SatHealth, a multimodal public dataset in Ohio consisting of environmental variables, satellite images, SDoH, and regional disease prevalence. To the best of our knowledge, SatHealth is the first dataset in the US that combines regional environmental characteristics with a healthcare database. We designed an embedding pipeline to fuse multimodal environmental data and produce regional living environment representations.
Using the representation, we conduct experiments on two tasks: regional public health modeling and personalized disease risk prediction. The experimental results show that multimodal environmental information helps boost model performance in these problems. We also find that leveraging spatiotemporal information helps improve model robustness. Finally, we deployed a web application for users to explore and download our dataset.

\begin{acks}
We acknowledge the funding for the project provided by the National Institute of Allergy and Infectious Diseases (R01AI188576), the National Institute on Drug Abuse (R01DA057668), and the National Science Foundation (2145625).
\end{acks}

\bibliographystyle{ACM-Reference-Format}
\bibliography{references}

\appendix


\section{Dataset Information}

\subsection{Data Scope}

We started with data curation in Ohio while participating in the Ohio O-SUDDEn program~\cite{osudden_proj}. We noticed the shortage of medical datasets with environmental health risk factors. Therefore, we develop SatHealth, a multimodal public health dataset with environmental factors in multiple categories. We use the data from 2016 to 2022 that is available for all modalities. 

We show the scope of SatHealth in Table \ref{tab:supp_data_scope}. SatHealth includes monthly aggregated data for climate, air quality, and greenery index at multiple geographical area levels, including counties, ZIP Code Tabulation Areas (ZCTAs), census tracts, and Core Based Statistical Areas (CBSAs). Land cover data in SatHealth is time-independent and denotes the area fraction of multiple land usage types at multiple geographical area levels mentioned above. Moreover, our dataset contains the latest aerial satellite images requested from Google Maps~\cite{google_static_map}; each image covers a square area around 500m wide. We've paid around 2500\$ (including free credits) to extract satellite images from the Google Maps API.

For health-related outcomes, we have the Social Deprivation Index (SDI) in 2019. We also provide yearly prevalences for all diseases identified by ICD codes estimated from medical claims in Ohio extracted from the MarketScan database~\cite{merative_real_2024}. We also paid for authorized access to the MarketScan database.

Although SatHealth was created in Ohio using data from 2016-2022, our framework is scalable, and we will gradually cover the whole US and keep updating new data after 2022. We also provide the code for users to create their customer data. 

\subsection{Ethics and Fairness Statement}
We list the source and data usage license for components in our dataset in Table \ref{tab:supp_data_license}.

\begin{table}[t]
\centering
\caption{SatHealth Data Scope}
\label{tab:supp_data_scope}
\begin{tabular}{lll}
\toprule
                             & Spatial Range                            & Time Span           \\\midrule
SatHealth                    & Ohio                                     & 2016-2022           \\\bottomrule
                             & Spatial Res.                       & Temporal Res. \\\midrule
Climate/Air Q/Greenery & Cty/ZCTA/CT/CBSA             & Month               \\
Land cover                   & Cty/ZCTA/CT/CBSA             & N/A                 \\
SDoH (SDI score)             & Cty/ZCTA/CT             & Year                \\
Disease Prevalence           & CBSA                                     & Year                \\
Satellite image              & 500m$\times$500m per image & N/A                \\\bottomrule
\multicolumn{3}{l}{\small Res.: Resolution; Air Q: Air quality; CBSA: Core Based Statistical Area}\\
\multicolumn{3}{l}{\small Cty: County; ZCTA: Zip Code Tabulation Area; CT: Census Tract}
\end{tabular}
\end{table}

\begin{table*}[ht]
\centering
\caption{SatHealth Data Source and License}
\label{tab:supp_data_license}
\begin{tabular}{lllll}
\toprule
Item               & Source       & Resolution & Frequency & License                                              \\\midrule
Climate/Greenery & \href{https://www.ecmwf.int/en/forecasts/dataset/ecmwf-reanalysis-v5}{ECMWF Climate Reanalysis (ERA5) }             & 11132 m & 1 month & \href{https://apps.ecmwf.int/datasets/licences/copernicus/}{link - Terms of Service} \\
Air Q - NO2      & \href{https://sentiwiki.copernicus.eu/web/sentinel-5p}{Sentinel-5P NRTI NO2}       & 1113.2 m & 2 days                & \href{https://sentinel.esa.int/documents/247904/690755/Sentinel\_Data\_Legal\_Notice}{link - Data Terms and Conditions}                       \\
Air Q - Ozone    & \href{https://developers.google.com/earth-engine/datasets/catalog/TOMS_MERGED}{TOMS and OMI Merged Ozone Data}             & 111000 m & 1 day                 & \href{https://www.earthdata.nasa.gov/engage/open-data-services-software-policies/data-information-guidance}{Data Information Guidance} \\
Air Q - Others   & \href{https://ads.atmosphere.copernicus.eu/datasets/cams-global-atmospheric-composition-forecasts?tab=overview}{Copernicus Atmosphere Monitoring Service (CAMS)}       & 44528 m & 1 day & \href{https://apps.ecmwf.int/datasets/licences/copernicus/}{link - Terms of Service}                                                 \\
Greenery - NDVI  & \href{https://sentiwiki.copernicus.eu/web/sentinel-2}{Harmonized Sentinel-2 MSI, Level-2A}        & 10 m & 5 days & \href{https://sentinel.esa.int/documents/247904/690755/Sentinel\_Data\_Legal\_Notice}{link - Data Terms and Conditions}                       \\
Land cover             & \href{https://land.copernicus.eu/en/products/global-dynamic-land-cover}{The Copernicus Global Land Service (CGLS)} & 100 m & 1 year & free and open to all users                   \\
SDI score              & \href{https://www.graham-center.org/maps-data-tools/social-deprivation-index.html}{Social Deprivation Index (SDI)}                       & - & -  & publicly available                                     \\
MarketScan CCAE            & \href{https://www.merative.com/documents/merative-marketscan-research-databases}{Merative MarketScan Research Databases}              & - & -                  & private                                              \\
Satellite image        & \href{https://developers.google.com/maps/documentation/maps-static/overview}{Google Static Map}                                     & - & - & \href{https://cloud.google.com/maps-platform/terms}{link - Terms of Service}  \\\bottomrule  
\multicolumn{5}{l}{Resolution: pixel size when mapped to geographical area; Frequency: time period between records}
\end{tabular}
\end{table*}

The MarketScan database used in this research is fully Health HIPAA compliant, de-identified, and has very minimal risk of the potential for loss of privacy. We will only publish regional statistics from the original medical claims and exclude regions with fewer than 10 people. Thus, there's a minimal privacy leakage issue for our medical data. 

\subsection{Data License}
The dataset is released under the CC BY-SA 4.0 license.

\subsection{Data Access}
\label{sec:supp_data_access}
Users can explore and download our dataset at \url{https://aimed-sathealth.net} with documentation at \url{https://github.com/Wang-Yuanlong/SatHealth}. We provide monthly data for climate, air quality, and greenery of multiple cartographic levels mentioned above for users to download. We also offer regional SDI scores and disease prevalence for all ICD codes in the MarketScan database. However, personalized data is not available. Moreover, the collection of satellite images from Google Maps is not available due to copyright concerns, but we provide scripts and a manual for users to extract them from the API Google provided.

\section{Dataset Details}

\subsection{File Structure}
Our dataset and embeddings can be found at \url{https://aimed-sathealth.net}. We display the downloaded file structure in Table \ref{tab:supp_data_structure}.

\begin{table*}[t]
\centering
\caption{SatHealth File Structure. Curly bracket (\{\}) means there are copies of the same structure for each item in the bracket.}
\label{tab:supp_data_structure}
\resizebox{0.98\textwidth}{!}{
\begin{tabular}{lll}
\toprule
\multicolumn{3}{c}{Raw data}                                                                                                                                  \\\midrule
File                              & Primary Key                         & Note                                                                                \\\midrule
\{County, CBSA, ZCTA, CT\}/       &                                     & 4 folders for different geo-regions                                                 \\
\;+ climate.csv                     & (GEOID, year, month)                &                                                                                     \\
\;+ greenery.csv                    & (GEOID, year, month)                &                                                                                     \\
\;+ airquality.csv                  & (GEOID, year, month)                &                                                                                     \\
\;+ landcover.csv                   & GEOID                               &                                                                                     \\
\;+ sdi.csv                         & GEOID                               & Ohio subset of the 2019 SDI downloaded from the \href{https://www.graham-center.org/maps-data-tools/social-deprivation-index.html}{SDI website}, no CBSA version \\
\;+ google\_map\_points\_linked.csv & (GEOID, county, row\_idx, col\_idx) & Mapping from regions to locations in google\_map\_points.csv to cover the regions            \\
icd\{l1,l2,l3\}\_prev\_ohio.csv   & (GEOID, year, code)                 &                                                                                     \\
column\_dictionary.csv            & (table, column)                     & A collection of column description in all CSV files                                     \\
google\_map\_request.py           & -                                   & Scripts for requesting satellite images from Google Static Map API given locations  \\
google\_map\_points.csv           & (county, row\_idx, col\_idx)        & A collection of locations used for requesting satellite images                      \\
README.md                         & -                                   & Documentation                                                                       \\\toprule
\multicolumn{3}{c}{Embedding}                                                                                                                                 \\\midrule
File                              & Primary Key                         & Note                                                                                \\\midrule
\{County, CBSA, ZCTA, CT\}/       &                                     & 4 folders for different geo-regions                                                 \\
\;+ embedding.csv                   & (GEOID, year)                       &                                                                                     \\
\;+ embedding\_T\_agg.csv           & (GEOID, year)                       & Regional environment embeddings enhanced by history embeddings                      \\
\;+ embedding\_S\_agg.csv           & (GEOID, year)                       & Regional environment embeddings enhanced by neighborhood embeddings                 \\
README.md                         & -                                   & Documentation                    \\\bottomrule                                                  
\end{tabular}}
\end{table*}

\subsection{Dataset Components}

SatHealth contains multimodal data from multiple sources.

\subsubsection{Environmental Data}

\textbf{ERA5-Land.} ERA5-Land is a global reanalysis dataset developed by the European Centre for Medium-Range Weather Forecasts (ECMWF) under the Copernicus Climate Change Service (C3S). It offers hourly data on land surface variables from 1950 to 5 days before the current date at a resolution of nearly 9km. By focusing on land components, ERA5-Land provides a consistent and detailed depiction of the water and energy cycles over land, making it a valuable resource for hydrological studies, numerical weather prediction, and environmental management applications. Our study uses the monthly aggregated version with a subset of variables following~\cite{scepanovic2024medsat}. We conduct a regional average reduction of all variables on all cartographic levels we presented before. 

\textbf{Global Dynamic Land Cover.} The Copernicus Global Land Monitoring Service offers the Global Dynamic Land Cover product, providing annual global land cover maps and cover fraction layers at a 100-meter resolution. These maps are classified into 23 discrete categories based on the UN-FAO Land Cover Classification System, offering detailed insights into Earth's surface composition. Additionally, the product includes continuous field layers, or "fraction maps," which estimate the proportional coverage of various land cover types within each pixel, enhancing the representation of heterogeneous areas. Our study incorporates the land cover fraction of 9 land cover types following~\cite{scepanovic2024medsat}. We conduct a regional average reduction of the cover fractions on all cartographic levels we presented before.

\subsubsection{Satellite Image}
Google Maps and Google Earth use satellite imagery and aerial photography to provide detailed, up-to-date maps. The imagery is sourced from various providers, including high-resolution data from companies like Maxar Technologies (formerly DigitalGlobe), and is processed to create seamless visual representations of Earth's surface. The Google Maps Static API~\cite{google_static_map} allows developers to embed these images as static, non-interactive maps into applications via simple HTTP requests, specifying parameters like location, zoom level, size, and markers. This approach offers a lightweight alternative to getting high-quality satellite images without cloud interference at arbitrary locations. We use Google Static Maps API to query images at zoom level 17 to cover Ohio state, which results in 432,918 images. Note that images from Google do not have specific timestamps and may not be well-synchronized. However, this doesn't matter if we assume the stability of the landscape in several years.

\subsubsection{Social Determinants of Health}
The Social Deprivation Index (SDI), developed by the Robert Graham Center, is a composite measure designed to quantify levels of social disadvantage across small geographic areas in the United States~\cite{sdi_def}. Utilizing data from the American Community Survey, the SDI incorporates seven demographic characteristics: percentage of individuals in poverty, adults with less than a high school education, single-parent households, rented housing units, overcrowded housing conditions, households without a vehicle, and non-employed adults under 65. SDI converts these census statistics into centiles and combines them to produce the final SDI score. Our dataset includes the county, zip code tabulation area, and census tract-level SDI score in Ohio.

\subsubsection{Health Outcomes}
The IBM MarketScan Commercial Claims and Encounters (CCAE) Database is a comprehensive resource that compiles de-identified healthcare claims data from over 250 million individuals in the United States. This database encompasses information from the medical experience of insured employees and their dependents for active employees, early retirees, Consolidated Omnibus Budget Reconciliation Act (COBRA) continues, and Medicare-eligible retirees with employer-provided Medicare Supplemental plans. It integrates detailed records of inpatient and outpatient services, prescription drug claims, and enrollment data, offering a longitudinal perspective on healthcare utilization and expenditures. Our dataset includes 2141777 patients with Ohio residency. We estimated the disease prevalence based on the ICD code occurrence in the dataset.

We create embeddings based on the collected data according to the process described in Section \ref{sec:data_construction}. A complete list of variables and image features covered in SatHealth can be found in Table \ref{tab:supp_data_features}.

\section{Additional Experimental Details and Results}
We put the additional experimental results in this section.

\subsection{Spatiotemporal Generalizability Analysis}
\label{sec:ts_gen}

In public health studies, a model's temporal and spatial generalizability is essential as we want our models to be robust as time goes by or when we apply them to other places~\cite{tesema_systematic_2023}. Therefore, we conduct additional experiments under spatiotemporal distribution shift cases and provide guidelines for users to develop robust models. All experiments conducted in this section are based on the regional public health modeling task as in Section \ref{sec:regional_public_health_modeling}.

\subsubsection{Temporal-spatial Generalization Scenarios}
\label{sec:ts_cv}
We present three scenarios of spatiotemporal generalization settings: spatial interpolation \& extrapolation~\cite{roberts_cross-validation_2017, goodchild_replication_2021}, and temporal forecasting. We simulate the scenarios with SatHealth by designing different cross-validation strategies.

\textbf{Spatial interpolation} measures the model ability of within-area prediction~\cite{sun_spatial_2023}. In this setting, regions that are adjacent or close to each other can be dispatched to different folds. Hence, models can leverage similarity between adjacent areas and generalize to adjacent areas. We implement this by randomly splitting regions into K-folds; each fold contains data from all years of its regions.

\textbf{Spatial extrapolation} measures the model ability of between-area prediction~\cite{sun_spatial_2023}. In this setting, models are trained in some region clusters and tested in some distant clusters. Therefore, spatial-agnostic patterns captured by the model are evaluated. We implement this by clustering regions by their centroid using K-means and using K clusters as K-folds. 

\textbf{Temporal forecasting} measures the model's ability to forecast future outcomes from historical records. This is implemented using the first 4 years (2016-2019) for training and the remaining 3 years (2020-2022) for testing.

\subsubsection{Spatiotemporal Enhanced Regression} 
\label{sec:ts_regression}

Our dataset can be formalized as $\mathcal{D}=\{X^s_{r},X^d_{r,t},Y_{r,t}\}_{r\in R,t\in T}$. Here, $R$ and $T$ represent the regions and the timestamps, respectively. $X^s_r$ denotes static, time-invariant features. For example, we regard satellite images and land cover as static features, as they remain relatively stable over our study period. $X^d_{r,t}$ denotes time-dependent features, including air quality, climate, and greenery variables. $Y_{r,t}$ corresponds to health outcomes for each region and year, including SDoH or disease prevalence.

As in Section \ref{sec:regional_public_health_modeling}, we fit regression models to predict regional SDI and ICD code prevalence by regional environmental representations. We start with the trivial baseline model without any spatiotemporal effect considered. Formally, the baseline regression model can be expressed as:

\begin{align*}
        Y_{r,t} &= \hat{Y}_{r,t} + \epsilon\\
        \hat{Y}_{r,t} &= f(X^s_r, X^d_{r,t})
\end{align*}

$\hat{Y}_{r,t}$ is the estimated outcome, $f$ is the regression model, and $\epsilon$ is the unmeasured variation.

Intuitively, regional health outcomes have inherent spatial and temporal autocorrelation because of the similar environment of adjacent regions and gradual temporal shift. Therefore, regression models could benefit from leveraging neighborhood and historical information. Inspired by the boosting strategy for ensemble learning, we introduce the spatiotemporal correlation to the baseline regression model:

\begin{align*}
    \hat{Y}_{r,t}&=f(X^s_r, X^d_{r,t})\\
    &+ g_s(\{(X^s_n,X^d_{n,t})\;|\;n\in N_r\}) + g_t(\{X^d_{r,h}|h<t\})
\end{align*}

$X^s_n, X^d_{n,t}$ are environment factors from neighborhood regions $N_r$ of the region $r$; $X^d_{r,h}$ are temporal features from history records of the region $r$. $g_s,g_t$ are models learned to account for spatial and temporal information. In our implementation, we consider environmental factors from neighborhoods or history. It will first aggregate the neighborhood and history input features using a weighted sum with an exponential decay factor for history aggregation and constant coefficients for neighborhood aggregation. Then, it fits two regressors for neighborhood and temporal input vectors.

The enhanced model is trained in a boosting manner. We first train $f$ with the original features and outcomes. Then, $g_s$ is trained to predict the residual of $f$, and $g_t$ is trained to predict the residual of $f+g_s$.

\subsection{Other Experimental Results}

Table \ref{tab:supp_ur-test} shows the level-3 ICD codes with the highest prevalence disparity by t-test or odds ratio. It can be seen that the odds ratio results align with Table \ref{tab:ur-test}.

\begin{table*}[ht]
    \centering
    \caption{ICD codes with top Urban-rural prevalence disparity by t-test and Odds Ratio}
    \label{tab:supp_ur-test}
    \begin{tabular}{cllll}
    \toprule
    \multicolumn{5}{c}{\textbf{t-Test}}\\\midrule
\multicolumn{1}{l}{Category}     & ICD code & t-statistics & p-value & Description                                                                           \\\midrule
\multirow{5}{*}{Urban Prevalent} & H02      & 4.983        & 0.0003  & Other disorders of eyelid                                                             \\
                                 & H00      & 5.174        & 0.0005  & Hordeolum and chalazion                                                               \\
                                 & L42      & 4.636        & 0.0006  & Pityriasis rosea                                                                      \\
                                 & I78      & 4.487        & 0.0007  & Diseases of capillaries                                                               \\
                                 & L80      & 4.353        & 0.0038  & Vitiligo                                                                              \\\midrule
\multirow{5}{*}{Rural Prevalent} & E16      & -6.725       & 0.0000  & Other disorders of pancreatic internal secretion                                      \\
                                 & K21      & -6.469       & 0.0000  & Gastro-esophageal reflux disease                                                      \\
                                 & R01      & -5.577       & 0.0002  & Cardiac murmurs and other cardiac sounds                                              \\
                                 & Z90      & -5.290       & 0.0002  & Acquired absence of organs                                                            \\
                                 & N99      & -4.962       & 0.0003  & \makecell[l]{Intraoperative and postprocedural complications\\and disorders of genitourinary system}\\\bottomrule
\multicolumn{5}{c}{\textbf{Odds Ratio}}\\\midrule
\multicolumn{1}{l}{Category}     & ICD code & Odds-ratio & 95\% CI        & Description                                                                        \\\midrule
\multirow{5}{*}{Urban Prevalent} & P01      & 4.282     & (3.431, 5.343) & Newborn affected by maternal complications of pregnancy                            \\
                                 & P03      & 2.431     & (2.115, 2.795) & Newborn affected by other complications of labor and delivery                      \\
                                 & P00      & 2.081     & (1.799, 2.407) & \makecell[l]{Newborn affected by maternal conditions that may be\\unrelated to present pregnancy} \\
                                 & P54      & 1.945     & (1.358, 2.784) & Other neonatal hemorrhages                                                         \\
                                 & P12      & 1.863     & (1.452, 2.389) & Birth injury to scalp                                                              \\\midrule
\multirow{5}{*}{Rural Prevalent} & H68      & 0.283     & (0.237, 0.336) & Eustachian salpingitis and obstruction                                             \\
                                 & E02      & 0.371     & (0.275, 0.502) & Subclinical iodine-deficiency hypothyroidism                                       \\
                                 & N42      & 0.385     & (0.330, 0.449) & Other and unspecified disorders of prostate                                        \\
                                 & D34      & 0.387     & (0.298, 0.503) & Benign neoplasm of thyroid gland                                                   \\
                                 & D13      & 0.407     & (0.354, 0.469) & Benign neoplasm of other and ill-defined parts of digestive system \\\bottomrule
\end{tabular}
\end{table*}

\begin{figure*}[ht]
    \centering
    \includegraphics[width=0.8\textwidth]{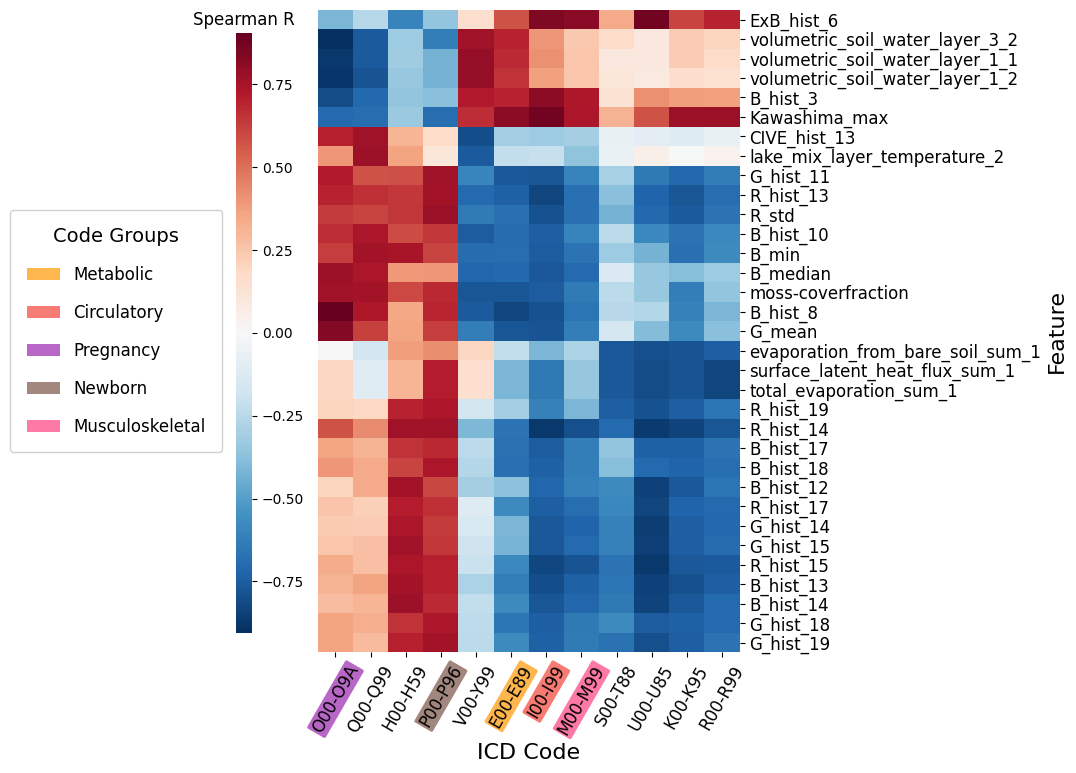}
    \caption{Feature correlations - top level ICD}
    \label{fig:feat-corr1}
\end{figure*}

\begin{figure*}[ht]
    \centering
    \includegraphics[width=0.95\textwidth]{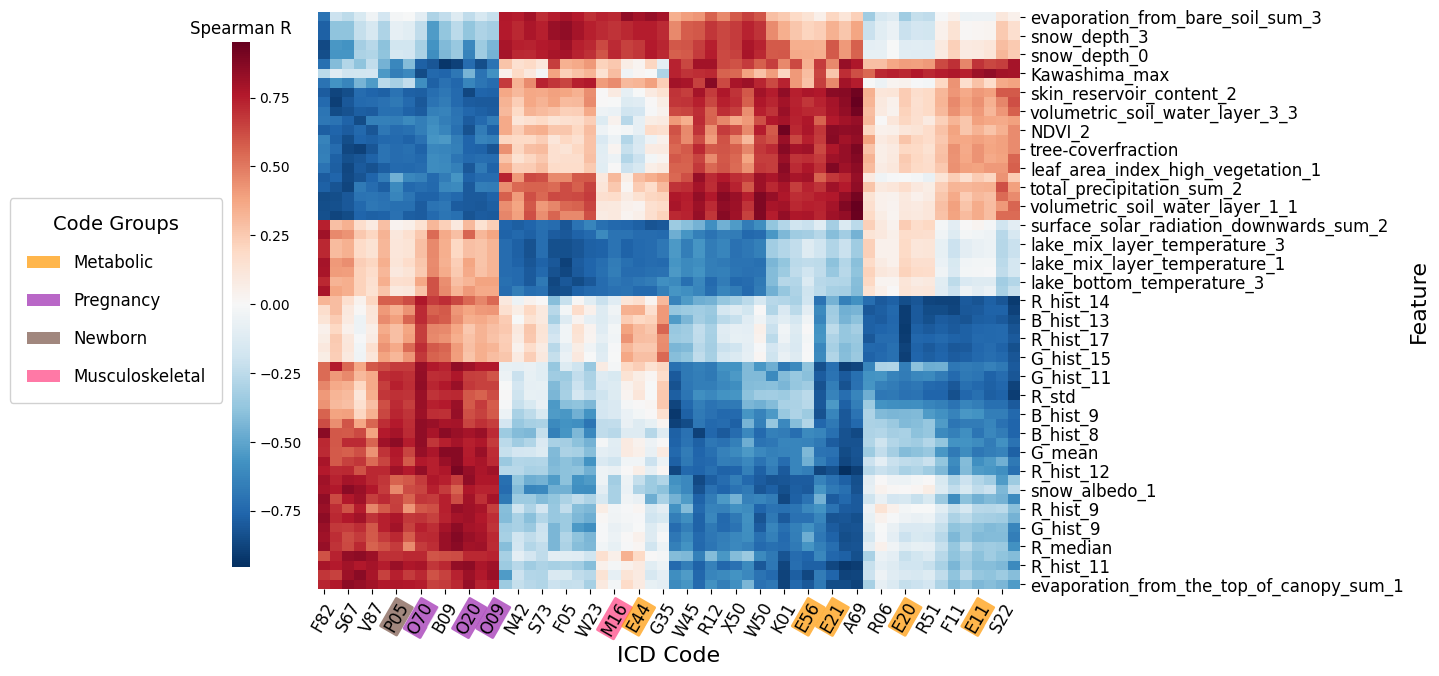}
    \caption{Feature correlations - third level ICD}
    \label{fig:feat-corr}
\end{figure*}

We display the feature-disease correlation for top-level and level-3 ICD codes in Figure \ref{fig:feat-corr1} and \ref{fig:feat-corr}. The results also show clustering characteristics and similar correlation patterns to Figure \ref{fig:feat-corr2}.

\begin{table*}[t]
    \centering
    \caption{SatHealth Variables and Features. Curly bracket (\{\}) means item copies for each item in the bracket.}
    \label{tab:supp_data_features}
    \resizebox{!}{\textheight}{
    \Rotatebox{90}{
    \begin{tabular}{cll}
    \toprule
Category                                  & Variable/Feature                                                 & Description                                                                                                                                  \\\midrule
\multirow{20}{*}{Climate}                 & dewpoint\_temperature\_2m                                        & Temperature to which the air, at 2 meters above the surface of the Earth, would have to be cooled for saturation to occur.                   \\
                                          & temperature\_2m                                                  & Temperature of air at 2m above the surface of land, sea or in-land waters.                                                                   \\
                                          & soil\_temperature\_level\_\{1, 3\}                               & Temperature of the soil in different soil layer of the ECMWF Integrated Forecasting System                                                   \\
                                          & lake\_bottom\_temperature                                        & Temperature of water at the bottom of inland water bodies (lakes, reservoirs, rivers) and coastal waters.                                    \\
                                          & lake\_mix\_layer\_\{depth, temprature\}                          & The thickness/temprature of the upper most layer of an inland water body (lake, reservoirs, and rivers) or coastal waters that is well mixed \\
                                          & lake\_total\_layer\_temperature                                  & The mean temperature of total water column in inland water bodies (lakes, reservoirs and rivers) and coastal waters.                         \\
                                          & snow\_\{albedo, cover, density, depth\}                          & Multiple measurements of snow layer in the ECMWF Integrated Forecast System                                                                  \\
                                          & skin\_reservoir\_content                                         & Amount of water in the vegetation canopy and/or in a thin layer on the soil.                                                                 \\
                                          & volumetric\_soil\_water\_layer\_\{1, 3\}                         & Volume of water in in different soil layer of the ECMWF Integrated Forecasting System.                                                       \\
                                          & surface\_latent\_heat\_flux\_sum                                 & Exchange of latent heat with the surface through turbulent diffusion.                                                                        \\
                                          & surface\_net\_solar\_radiation\_sum                              & Amount of solar radiation reaching the surface of the Earth minus the amount reflected by the Earth's surface.                               \\
                                          & surface\_\{solar, thermal\}\_radiation\_downwards\_sum           & Amount of solar radiation (also known as shortwave radiation) reaching the surface of the Earth.                                             \\
                                          & evaporation\_from\_bare\_soil\_sum                               & Amount of thermal (also known as longwave or terrestrial) radiation emitted by the atmosphere and clouds that reaches the Earth's surface.   \\
                                          & evaporation\_from\_the\_top\_of\_canopy\_sum                     & The amount of evaporation from the canopy interception reservoir at the top of the canopy.                                                   \\
                                          & evaporation\_from\_open\_water\_surfaces\_excluding\_oceans\_sum & Amount of evaporation from surface water storage like lakes and inundated areas but excluding oceans.                                        \\
                                          & total\_evaporation\_sum                                          & Accumulated amount of water that has evaporated from the Earth's surface into vapor in the air above.                                        \\
                                          & \{u, v\}\_component\_of\_wind\_10m                               & Eastward/Northward component of the 10m wind.                                                                                                \\
                                          & surface\_pressure                                                & Pressure (force per unit area) of the atmosphere on the surface of land, sea and in-land water.                                              \\
                                          & total\_precipitation\_sum                                        & Accumulated liquid and frozen water, including rain and snow, that falls to the Earth's surface.                                             \\
                                          & surface\_runoff\_sum                                             & Total amount of water accumulated from the beginning of the forecast time to the end of the forecast step                                    \\\midrule
\multirow{3}{*}{Greenery}                 & leaf\_area\_index\_\{high, low\}\_vegetation                     & One-half of the total green leaf area per unit horizontal ground surface area for high/low vegetation type.                                  \\
                                          & NDVI                                                             & NDVI index calculated from multispectual image in the area                                                                                   \\
                                          & NDVI\_binary                                                     & Percentage of pixels with NDVI index above 0.2 in the area                                                                                   \\\midrule
\multirow{4}{*}{Air Quality}              & total\_aerosol\_optical\_depth\_at\_550nm\_surface               & Total Aerosol Optical Depth at 550 nm                                                                                                        \\
                                          & particulate\_matter\_d\_less\_than\_25\_um\_surface              & Concentration of particulate matter d \textless 2.5 um                                                                                       \\
                                          & NO2\_column\_number\_density                                     & Total vertical column of NO2 (ratio of the slant column density of NO2 and the total air mass factor).                                       \\
                                          & ozone                                                            & Total column ozone                                                                                                                           \\\midrule
\multirow{10}{*}{RGB Image Indices}       & \{R,G,B\}                                                        & Pixel value of Red, Green, and Blue channel. Will use r, g, b for normalized pixel value {[}r=R/(R+G+B), etc.{]} below.                      \\
                                          & Excess Green Index (ExG)                                         & 2 * g - r - b                                                                                                                                \\
                                          & Excess Red Index (ExR)                                           & 1.4 * r - g                                                                                                                                  \\
                                          & Excess Blue Index (ExB)                                          & 1.4 * b - g                                                                                                                                  \\
                                          & Green Red Vegetation Index (GRVI)                                & (G-R)/(G+R)                                                                                                                                  \\
                                          & Modified Green Red Vegetation Index (MGRVI)                      & (G\textasciicircum{}2-R\textasciicircum{}2)/(G\textasciicircum{}2+R\textasciicircum{}2)                                                      \\
                                          & Red Green Blue Vegetation Index (RGBVI)                          & (G\textasciicircum{}2-B*R)/(G\textasciicircum{}2+B*R)                                                                                        \\
                                          & Kawashima Index                                                  & (R-B)/(R+B)                                                                                                                                  \\
                                          & Color index of vegetation extraction (CIVE)                      & 0.441r - 0.811g + 0.385b + 18.78745                                                                                                          \\
                                          & Green leaf index (GLI)                                           & (2g-r-b)/(2g+r+b)                                                                                                                            \\\midrule
\multirow{2}{*}{\makecell[c]{Image Embedding\\Features}} & \textless{}index\textgreater{}\_\{max, min, median, mean, std\}  & Statistics among all pixels in an image for each index in the above section as \textless{}index\textgreater{}                                \\
                                          & \textless{}index\textgreater{}\_hist\_\{0…19\}                   & Number of pixels with index value in histogram bin 0$\sim$19 for each index in the above section as \textless{}index\textgreater{}     \\\bottomrule     
\end{tabular}}}
\end{table*}

\end{document}